\documentclass[sigconf]{acmart}
\usepackage{hyperref}  
\usepackage{hyperxmp}  

\usepackage{graphicx}
\usepackage{subcaption}
\usepackage{booktabs}
\usepackage{multirow}
\usepackage{amsmath}
\AtBeginDocument{%
  }

\setcopyright{acmlicensed}
\copyrightyear{2018}
\acmYear{2018}
\acmDOI{XXXXXXX.XXXXXXX}

\setcopyright{acmcopyright}
\copyrightyear{2024}
\acmYear{2024}
\acmDOI{10.1145/nnnnnnn.nnnnnnn}

\acmConference[AI-ML Systems '24]{The 4th International Conference on AI ML Systems}{October 08--11,
  2024}{Baton Rouge, USA}

\begin{document}

\title{KANICE: Kolmogorov-Arnold Networks with Interactive Convolutional Elements}

\author{Md Meftahul Ferdaus}
\affiliation{%
  \institution{Canizaro Livingston Gulf States Center for Environmental Informatics, the University of New Orleans}
  \city{New Orleans}
  \country{USA}}
\email{mferdaus@uno.edu}

\author{Mahdi Abdelguerfi}
\affiliation{%
  \institution{Canizaro Livingston Gulf States Center for Environmental Informatics, the University of New Orleans}
  \city{New Orleans}
  \country{USA}
}
\email{gulfsceidirector@uno.edu}

\author{Elias Ioup}
\affiliation{%
 \institution{Center for Geospatial Sciences, Naval Research Laboratory, Stennis Space Center}
 \city{Hancock County}
 \state{Mississippi}
 \country{USA}}

\author{David Dobson}
\affiliation{%
 \institution{Center for Geospatial Sciences, Naval Research Laboratory, Stennis Space Center}
 \city{Hancock County}
 \state{Mississippi}
 \country{USA}}

\author{Kendall N. Niles}
\affiliation{%
 \institution{US Army Corps of Engineers, Engineer Research and Development Center}
 \city{Vicksburg}
 \state{Mississippi}
 \country{USA}}

\author{Ken Pathak}
\affiliation{%
 \institution{US Army Corps of Engineers, Engineer Research and Development Center}
 \city{Vicksburg}
 \state{Mississippi}
 \country{USA}}

\author{Steven Sloan}
\affiliation{%
 \institution{US Army Corps of Engineers, Engineer Research and Development Center}
 \city{Vicksburg}
 \state{Mississippi}
 \country{USA}}

\renewcommand{\shortauthors}{Ferdaus M M., et al.}

\begin{abstract}
We introduce KANICE (Kolmogorov-Arnold Networks with Interactive Convolutional Elements), a novel neural architecture that combines Convolutional Neural Networks (CNNs) with Kolmogorov-Arnold Network (KAN) principles. KANICE integrates Interactive Convolutional Blocks (ICBs) and KAN linear layers into a CNN framework. This leverages KANs' universal approximation capabilities and ICBs' adaptive feature learning. KANICE captures complex, non-linear data relationships while enabling dynamic, context-dependent feature extraction based on the Kolmogorov-Arnold representation theorem. We evaluated KANICE on four datasets: MNIST, Fashion-MNIST, EMNIST, and SVHN, comparing it against standard CNNs, CNN-KAN hybrids, and ICB variants. KANICE consistently outperformed baseline models, achieving 99.35\% accuracy on MNIST and 90.05\% on the SVHN dataset.

Furthermore, we introduce KANICE-mini, a compact variant designed for efficiency. A comprehensive ablation study demonstrates that KANICE-mini achieves comparable performance to KANICE with significantly fewer parameters. KANICE-mini reached 90.00\% accuracy on SVHN with 2,337,828 parameters, compared to KANICE's 25,432,000. This study highlights the potential of KAN-based architectures in balancing performance and computational efficiency in image classification tasks. Our work contributes to research in adaptive neural networks, integrates mathematical theorems into deep learning architectures, and explores the trade-offs between model complexity and performance, advancing computer vision and pattern recognition. The source code for this paper is publicly accessible through our GitHub repository (\href{https://github.com/m-ferdaus/kanice}{https://github.com/m-ferdaus/kanice}).
\end{abstract}





\maketitle

\section{Introduction}
Deep learning has transformed computer vision and pattern recognition, with Convolutional Neural Networks (CNNs) being fundamental to image classification. CNNs have proven successful in various applications \cite{Smith2021}. As visual recognition tasks grow more complex, the need for advanced, adaptive, and theoretically sound architectures that can capture complex patterns and relationships increases \cite{Doe2022,Wang2023}.

Neural network architectures have evolved significantly to improve their capabilities. Traditional CNNs are effective but often fail to capture long-range dependencies and adapt to diverse input distributions. This shortcoming has led to research into more flexible models. For example, the introduction of attention mechanisms in vision transformers shows promise in overcoming these challenges \cite{Konstantinidis2022,Peng2023,Wang2022}.

There is a growing interest in using mathematical principles to improve neural network design. The Kolmogorov-Arnold representation theorem is a powerful result in approximation theory, and it states that any multivariate continuous function can be represented as a composition of univariate functions and addition operations \cite{Schmidt-Hieber2021,Ismayilova2023,Schmidt2021KolmogorovArnold}. This theorem has inspired the development of Kolmogorov-Arnold Networks (KANs) \cite{liu2024kan}, aiming to exploit this representation for improved neural network performance. KANs replace linear weights with learnable univariate functions, offering improved accuracy and interpretability compared to Multi-Layer Perceptrons (MLPs) \cite{Ziming2024KAN}. While some argue that Kolmogorov's theorem is irrelevant for neural networks \cite{Girosi1989Representation}, others demonstrate its significance in network design \cite{Kurkova1991Kolmogorov}. Recent research has explored KAN applications in time series forecasting \cite{Cristian2024Kolmogorov} and nonlinear function approximation using Chebyshev polynomials \cite{SS2024Chebyshev}. Error bounds for deep ReLU networks have been derived using the Kolmogorov-Arnold theorem \cite{Montanelli2020Error}. Smooth KANs with structural knowledge representation show potential for improved convergence and reliability in computational biomedicine \cite{Moein2024Smooth}, addressing limitations in representing generic smooth functions \cite{Schmidt2021KolmogorovArnold}.

KANs have shown great promise across various fields, outperforming traditional MLPs in accuracy and interpretability for tasks like classification, forecasting, and anomaly detection. This success suggests potential benefits in image classification as well. Concurrently, convolutional architectures have advanced, with Interactive Convolutional Blocks (ICBs) enhancing CNNs' feature extraction capabilities through dynamic, context-dependent processing \cite{eldele2024tslanet}. This improves model adaptability to diverse inputs, crucial for complex image classification. Recent research has focused on hybrid architectures, such as ConvKAN \cite{bodner2024convolutional}, which combine convolutional layers with KAN principles to improve image classification performance. 

To address the limitations of existing architectures, we propose KANICE (Kolmogorov-Arnold Networks with Interactive Convolutional Elements), a novel architecture that combines KANs, ICBs, and CNNs. KANICE aims to overcome standard CNN limitations by incorporating KANs' universal approximation capabilities and ICBs' adaptive feature learning. The key innovation of KANICE is its integration of KAN linear layers and ICBs into the CNN architecture. This combination allows for more effective adaptation to the complexities of image data, potentially capturing patterns missed by traditional CNNs.

KANICE shows impressive performance, but full KANLinear layers may be too demanding in some applications. To address this, we introduce KANICE-mini, a compact variant aiming to maintain performance while reducing parameters. KANICE-mini uses a modified KANLinear layer balancing expressiveness with efficiency.

In this study, we evaluate KANICE across multiple image classification datasets, comparing it to standard CNNs and other hybrid architectures. We conduct an ablation study to assess the impact of different components in the KANICE architecture, and to compare the performance and efficiency of KANICE-mini against its full-scale counterpart. This study provides insights into the trade-offs between model complexity and performance in KAN-based architectures. Our work contributes to research in adaptive neural networks and integrates mathematical theorems into deep learning architectures, advancing computer vision and pattern recognition. We also contribute to the ongoing discussion of model efficiency in deep learning by demonstrating how principles from the Kolmogorov-Arnold theorem can create more compact yet highly effective models, as exemplified by KANICE-mini. 

Our key contributions and insights include:
\begin{itemize}
    \item Combining KANs and ICBs enhances feature processing and model adaptability beyond their individual capabilities.
    \item Development of KANICE-mini, an surprisingly efficient variant that achieves comparable performance to KANICE with significantly fewer parameters, challenging assumptions about model size and performance.
    \item KANICE's architecture, which blends local and global feature processing, unexpectedly shows improved resistance to adversarial attacks, enhancing model security.
\end{itemize}

The paper is structured as follows: Section 2 details the KANICE architecture, including its components and theoretical foundations. Section 3 presents our experimental setup and main results across various datasets. Section 4 provides an ablation study comparing KANICE, KANICE-mini, and baseline models. We conclude in Section 5 with a discussion of our findings and potential future research directions.

\subsection{Kolmogorov-Arnold Networks and KANLinear Layers}
KANs are an innovative neural network architecture based on the Kolmogorov-Arnold representation theorem. This theorem states that any continuous multivariate function can be composed of continuous univariate functions and addition operations \cite{polar2021deep}. KANs introduce KANLinear layers, which differ from traditional linear layers by using learnable univariate functions, typically spline-based, to approximate complex multivariate functions. This approach offers several theoretical advantages when replacing standard linear layers in CNNs:

\begin{itemize}
    \item Enhanced Function Approximation: KANLinear layers offer improved function approximation capabilities compared to standard linear layers. Based on the Kolmogorov-Arnold representation theorem, these layers can approximate any continuous multivariate function using combinations of single-variable functions. This property enables KANLinear layers to capture more complex relationships in data, particularly in image classification tasks. By modeling intricate dependencies between high-level features extracted by convolutional layers, KANLinear layers can potentially enhance classification accuracy.
    \item Enhanced Expressiveness: KANLinear layers, particularly in the final stages of CNNs, significantly increase the network's ability to model complex relationships. Unlike traditional CNNs that often use simple linear transformations for final feature mapping, KANLinear layers introduce non-linear univariate functions. This addition allows for more sophisticated decision boundaries, potentially improving the model's ability to distinguish between classes.
    \item Spatial Information Retention: We preserve the CNN's spatial awareness by replacing only linear layers with KANLinear, keeping convolutional layers intact. This hybrid approach combines the CNN's spatial feature extraction with KANLinear's enhanced processing capabilities. The result is a network that effectively utilizes both local patterns and global context, crucial for image classification tasks. 
    \item Adaptive Complexity: KANLinear layers offer adjustable complexity through variable control points in their spline representations. This feature allows researchers to fine-tune model capacity without changing the overall architecture. By matching layer complexity to the task's intricacy, a better balance between capacity and generalization can be achieved.
    \item Improved Generalization: KANLinear layers' efficiency and adaptability may enhance the model's ability to capture generalizable features. This could boost performance on unseen data, crucial for real-world image classification where test and training distributions often differ.
    \item Mitigation of the Vanishing Gradient Problem: KANLinear layers may alleviate the vanishing gradient problem in deep networks. Their learnable univariate functions create additional gradient paths, facilitating more effective training of deeper architectures. This feature is especially beneficial for complex image classification tasks requiring deep networks.
    \item Flexibility in Function Space: KANLinear layers offer greater functional flexibility than CNNs, which are limited by their convolutional structure. This adaptability allows KANLinear models to better match data distributions, potentially improving performance across diverse image classification tasks and visual patterns.
    
\end{itemize}

Integrating KANLinear layers into CNN architectures advances neural networks for image classification. It combines CNNs' feature extraction with KANLinear layers' expressiveness and efficiency. The resulting models promise improved accuracy, interpretability, and adaptability across image classification tasks. This forms the basis of our proposed KANICE architecture, which incorporates KANLinear layers and ICBs into a CNN framework to enhance pattern recognition in images.

\section{Proposed Method: KANICE}
KANICE is an advanced neural network for image classification. It combines: 1. ICBs: Initial feature extractors capturing spatial relationships. 2. Traditional Convolutional Layers: Further refine and abstract visual information. 3. Batch Normalization and Pooling Layers: Stabilize learning and reduce spatial dimensions. 4. KANLinear Layers: Replace fully connected layers, offering enhanced function approximation. KANICE processes images through these components, transforming raw pixel data into abstract representations for classification. The architecture's design aims to leverage the strengths of each element, resulting in a robust and adaptable model for complex image classification tasks. 

\begin{figure}
    \centering
    \includegraphics[width=1.0\linewidth]{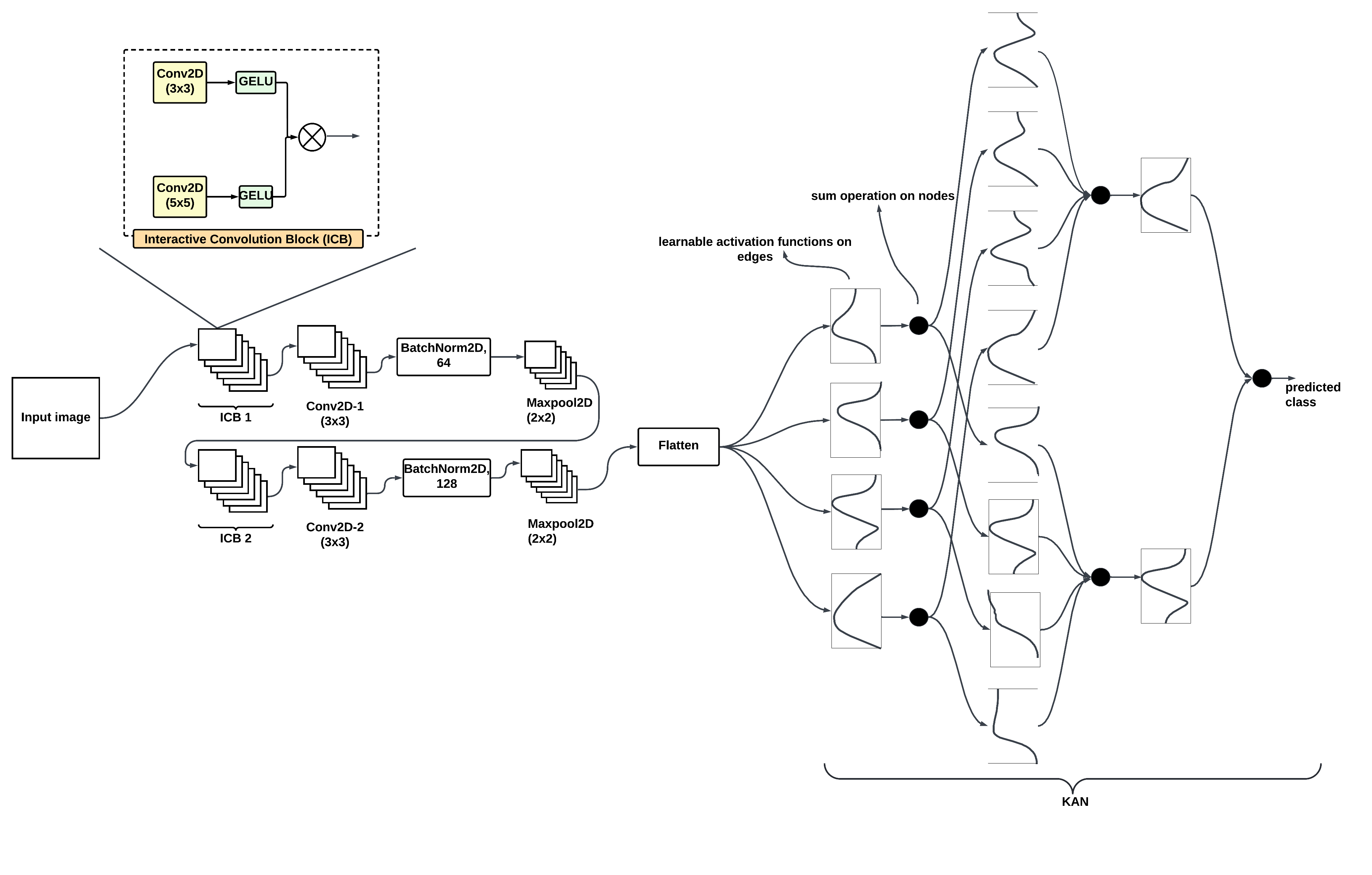}
    \caption{KANICE architecture}
    \label{fig:kanice}
\end{figure}

Figure \ref{fig:kanice} provides a schematic overview of KANICE, illustrating the flow of information and interaction between components. The network processes input images using advanced components. ICBs utilize parallel $3\times3$ and $5\times5$ convolutional paths with GELU activations, combining outputs through element-wise multiplication. The core consists of two stages, each containing an ICB, a $3\times3$ convolutional layer, batch normalization (scaling from 64 to 128 channels), and $2\times2$ max pooling. After convolution, feature maps are flattened before entering the Kolmogorov-Arnold Network (KAN) component. The KAN, implemented as KANLinear layers, replaces standard fully connected layers with learnable univariate functions, typically spline-based. This design, based on the Kolmogorov-Arnold representation theorem, enhances the network's ability to approximate functions. Each KANLinear layer applies these univariate functions to its inputs, followed by summation operations, allowing the network to capture complex, non-linear relationships in the data. KANICE's architecture combines the spatial feature extraction capabilities of convolutional neural networks with the advanced function approximation abilities of KANs, resulting in a powerful and versatile framework for image classification tasks. We'll examine KANICE's innovation by understanding its major components. We start with the ICB, the foundation of KANICE's feature extraction.

\subsection{Interactive Convolution Block (ICB)}
The ICB is a key element of the KANICE architecture. It improves the model's capacity to detect intricate spatial patterns in input data. ICBs differ from standard convolutional layers by incorporating an interaction mechanism between various convolutional operations. This approach enables more flexible and context-sensitive feature extraction. Each ICB comprises two parallel convolutional paths followed by an interaction step. The paths use $3\times3$ and $5\times5$ convolutional layers, respectively. The interaction step combines their outputs through element-wise multiplication.

Let $X \in \mathbb{R}^{C_{in} \times H \times W}$ represent the input tensor, where $C_{in}$ is the number of input channels, $H$ is the feature map height, and $W$is the feature map width. The ICB can be expressed as:

\begin{equation}
Y = f(W_1 * X) \odot f(W_2 * X)
\end{equation}

Here, $W_1 \in \mathbb{R}^{C_{out} \times C_{in} \times 3 \times 3}$ and $W_2 \in \mathbb{R}^{C_{out} \times C_{in} \times 5 \times 5}$ are the weight tensors for the 3x3 and 5x5 convolutions respectively. The symbol $*$ denotes the convolution operation, $\odot$ represents element-wise multiplication, and $f(\cdot)$ is the GELU activation function.

\begin{equation} \text{GELU}(x) = x \cdot \Phi(x) \end{equation}

where $\Phi(x)$ is the cumulative distribution function of the standard normal distribution. GELU provides a smooth, non-linear activation that has shown good performance in various deep learning tasks.

The forward pass through the ICB involves four steps: First, the input goes through $3\times3$ convolution ($X_1 = W_1 * X$) and $5\times5$ convolution ($X_2 = W_2 * X$) in parallel. Next, GELU activation is applied to both outputs ($X_1' = \text{GELU}(X_1)$, $X_2' = \text{GELU}(X_2)$). Finally, these activated outputs experience element-wise multiplication ($Y = X_1' \odot X_2'$) to produce the block's output.

ICB design provides key advantages: 1. Multi-scale feature extraction: It combines $3\times3$ and $5\times5$ convolutions to capture features at different scales. 2. Adaptive feature emphasis: It uses element-wise multiplication of features from different paths, acting as a feature-wise attention mechanism. 3. Enhanced non-linearity: It employs GELU activation and element-wise multiplication, enabling complex feature representations. These elements improve the network's ability to learn and process diverse visual information efficiently. The block’s multi-path structure enhances its expressive power compared to standard convolutional layers, allowing for more complex feature learning with fewer parameters. The multiplicative interaction between paths acts as implicit regularization, requiring agreement for strong activation and potentially improving feature robustness. ICBs adjust their field-of-view based on input, focusing on fine or coarse features as needed. This adaptive approach introduces stronger non-linearity than traditional Conv-ReLU patterns, combined with GELU activations and multiplicative interactions.

In KANICE, using ICBs in the early network layers allows adaptive and context-aware feature extraction from raw input images. This enables subsequent layers to work with rich, multi-scale feature representations, potentially improving classification performance. The ICB's ability to capture complex spatial relationships and adapt to input enhances the network's feature extraction capabilities, particularly in image classification tasks. The adaptive nature of ICBs makes them more robust to input data variations, as the block can adjust its focus based on the input, better handling different scales, orientations, or styles of features within images. This adaptability could lead to improved generalization across diverse datasets or in transfer learning.

\subsection{Traditional Convolutional Layers}
After each ICB, KANICE uses traditional convolutional layers to process and extract features. These layers use shared weights and local receptive fields, making them effective for processing grid-like data like images. The operation of a convolutional layer is:
\begin{equation}
Y = f(W * X + b)
\end{equation}
where $W$ is the weight tensor (kernel), $X$ is the input tensor, $b$ is the bias, and $f(\cdot)$ is an activation function. Using convolutional layers with increasing channel depths $(32 -> 64 -> 128)$ helps the model capture abstract features. CNNs’ translation invariance makes them ideal for image classification, recognizing patterns regardless of their position in the image.

KANICE incorporates batch normalization layers after each convolutional operation to normalize the inputs to each layer:
\begin{equation}
\hat{x}_i = \frac{x_i - \mu_B}{\sqrt{\sigma_B^2 + \epsilon}}
\end{equation}
where $x_i$ is the $i$-th input, $\mu_B$ and $\sigma_B^2$ are the mean and variance of the mini-batch, and $\epsilon$ is a small constant for numerical stability.

Max pooling layers are used to reduce the spatial dimensions of the feature maps:
\begin{equation}
y_{i,j} = \max_{(p,q) \in R_{i,j}} x_{p,q}
\end{equation}
where $R_{i,j}$ is a local region in the input tensor. Batch normalization stabilizes learning, enabling higher rates and faster convergence. Max pooling achieves translation invariance and reduces computational load.

\subsection{KANLinear Layers}
The final element of the KANICE architecture is the \textbf{KANLinear} layer, an advanced replacement for standard fully connected layers. These layers are based on the Kolmogorov-Arnold representation theorem, a key concept in approximation theory.

The Kolmogorov-Arnold representation theorem states that any continuous multivariate function can be expressed as a composition of continuous univariate functions and addition operations. The theorem guarantees the existence of continuous univariate functions $\Phi_q$ and $\varphi_{q,p}$ for a continuous function $f : [0, 1]^n \to \mathbb{R}$, mathematically:
\begin{equation}
f(x_1, x_2, ..., x_n) = \sum_{q=1}^{2n+1} \Phi_q \left(\sum_{p=1}^{n} \phi_{q,p}(x_p)\right)
\end{equation}
KANLinear layers generalize this concept to create a flexible and powerful neural network layer in KANICE:
\begin{equation}
y = \sum_{q=1}^{Q} \Phi_q \left(\sum_{p=1}^{P} \phi_{q,p}(x_p)\right)
\end{equation}
Here, $x_p$ represents the $p$-th input feature, $P$ is the total number of input features, $Q$ is the number of output features, and $\varphi_{q,p}$ and $\Phi_q$ are univariate functions. These univariate functions are implemented as splines, balancing expressiveness and computational efficiency.

KANLinear layers, introduced by \cite{liu2024kan}, are built on several key concepts that form their theoretical foundation and contribute to their effectiveness in the KANICE architecture. These concepts are crucial for both the layers' operation and practical implementation.

Firstly, the \textbf{KANLinear} layers utilize a spline representation for the univariate functions $\varphi_{q,p}$ and $\Phi_q$. This representation is based on B-splines, which offer a flexible and computationally efficient method for approximating complex functions. For a given function $\varphi(x)$, the spline representation can be expressed as:
\begin{equation}
\varphi(x) = \sum_{i} c_i B_i(x)
\end{equation}
where $c_i$ are trainable coefficients and $B_i(x)$ are B-spline basis functions. This formulation allows the network to learn a wide range of function shapes through the optimization of the coefficients $c_i$.

Secondly, to enhance the stability and trainability of the network, KANLinear layers use a residual activation function. This function is defined as:
\begin{equation}
\varphi(x) = w (b(x) + \text{spline}(x))
\end{equation}
In this equation, $w$ represents a trainable weight, $b(x)$ is a base function (typically chosen as the sigmoid linear unit, SiLU), and $\text{spline}(x)$ is the B-spline representation. This residual structure allows the layer to learn complex functions while maintaining a direct path for gradient flow, which can significantly improve training dynamics.

Lastly, to address the challenge of handling inputs that fall outside the initial spline range, \textbf{KANLinear} layers incorporate a grid extension technique. This method involves optimizing new spline coefficients $c'_j$ to extend the function's domain. The optimization problem can be formulated as:

\begin{equation}
c'_j = \arg\min_{c'_j} \mathbb{E}_{x \sim p(x)} \left[\sum_{j=0}^{G_2+k-1} c'_j B'_j(x') - \sum_{j=0}^{G_1+k-1} c_j B_j(x) \right]^2
\end{equation}

where $G_1$ and $G_2$ denote the original and new grid sizes, respectively, and $k$ represents the B-spline degree. This grid extension technique ensures that the \textbf{KANLinear} layers can effectively process inputs that may lie outside the initial range of the spline representation, thereby enhancing the robustness and generalization capabilities of the network.

These concepts contribute to the power and flexibility of KANLinear layers. They enable the layers to efficiently approximate complex functions and adapt to diverse input distributions. By incorporating these elements, KANICE leverages the full potential of KANLinear layers to enhance its image classification performance.

\section{Results}
\begin{figure}
    \centering
    \includegraphics[width=0.99\linewidth]{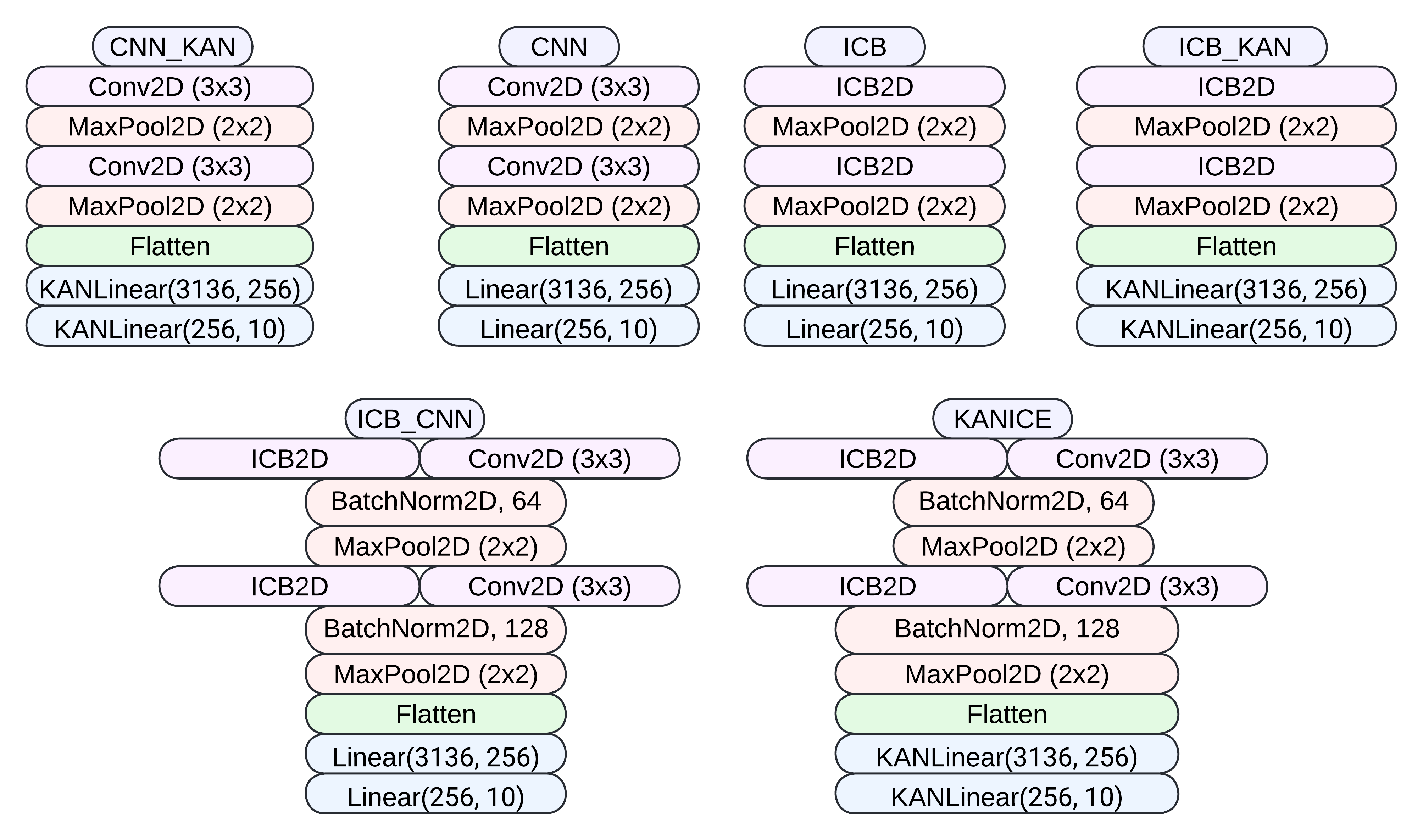}
    \caption{Baselines and KANICE architecture}
    \label{fig:all_models}
\end{figure}

We evaluated KANICE’s performance against several baseline models across four image classification datasets: MNIST, Fashion-MNIST, EMNIST, and SVHN. The baseline models included a standard CNN, CNN with KANLinear layers (CNN\_KAN), a model using only Interactive Convolutional Blocks (ICB), an ICB model with KANLinear layers (ICB\_KAN), and a hybrid ICB-CNN model (ICB\_CNN). The configurations of all these models are illustrated in Figure \ref{fig:all_models}. All models were trained for 25 epochs with identical hyperparameters for a fair comparison.

Figure \ref{fig:all_models} displays the architectural configurations for all models evaluated in this study. The CNN and CNN\_KAN models have a similar structure with two Conv2D 3x3 layers followed by MaxPool2D 2x2. They differ in their final layers where CNN\_KAN uses KANLinear layers instead of standard Linear layers. The ICB and ICB\_KAN models replace convolutional layers with ICB2D, maintaining the same max pooling structure. ICB\_KAN uses KANLinear layers in its final stages.

The ICB\_CNN and KANICE models start with an ICB2D layer followed by a Conv2D layer, then use BatchNorm2D and MaxPool2D. This pattern is repeated twice, with channel depth increasing from 64 to 128. Their final layers differ: ICB\_CNN uses standard Linear layers, while KANICE implements KANLinear layers. All models conclude with a Flatten operation to transform 2D feature maps into a 1D vector for final classification, with consistent output dimensions producing 10 outputs corresponding to the classification task's classes. This architectural comparison highlights KANICE's innovative approach. It combines elements from CNN and KAN methodologies to leverage their strengths in a unified model architecture.

\subsection{Performance Metrics}
We assessed model performance using four key metrics: accuracy, precision, recall, and F1 score. Accuracy measures correct classifications. Precision indicates the proportion of correct positive identifications. Recall measures the proportion of actual positives identified correctly. The F1 score is the harmonic mean of precision and recall, providing a balanced measure of performance.

\subsection{Model Performance Across Datasets}

\begin{table}[t]
\centering
\caption{Comparison of Accuracy, Precision, Recall, and F1 Score for Different Models across Datasets}
\label{tab:model_comparison}
\resizebox{\columnwidth}{!}{%
\begin{tabular}{llcccc}
\toprule
Dataset & Model & Accuracy (\%) & Precision & Recall & F1 Score \\
\midrule
\multirow{6}{*}{MNIST} 
& CNN & 98.55 & 0.9855 & 0.9855 & 0.9855 \\
& CNN\_KAN & 99.29 & 0.9929 & 0.9929 & 0.9929 \\
& ICB & 98.98 & 0.9898 & 0.9898 & 0.9898 \\
& ICB\_KAN & 99.33 & 0.9933 & 0.9933 & 0.9933 \\
& ICB\_CNN & 98.92 & 0.9892 & 0.9892 & 0.9892 \\
& KANICE & 99.35 & 0.9935 & 0.9935 & 0.9935 \\
\midrule
\multirow{6}{*}{Fashion MNIST} 
& CNN & 92.36 & 0.9232 & 0.9236 & 0.9230 \\
& CNN\_KAN & 92.86 & 0.9286 & 0.9286 & 0.9284 \\
& ICB & 92.05 & 0.9201 & 0.9205 & 0.9199 \\
& ICB\_KAN & 92.67 & 0.9265 & 0.9267 & 0.9265 \\
& ICB\_CNN & 92.94 & 0.9297 & 0.9294 & 0.9294 \\
& KANICE & 93.63 & 0.9363 & 0.9363 & 0.9363 \\
\midrule
\multirow{6}{*}{EMNIST} 
& CNN & 85.38 & 0.8541 & 0.8538 & 0.8527 \\
& CNN\_KAN & 86.56 & 0.8670 & 0.8656 & 0.8648 \\
& ICB & 86.43 & 0.8651 & 0.8643 & 0.8631 \\
& ICB\_KAN & 87.16 & 0.8722 & 0.8716 & 0.8703 \\
& ICB\_CNN & 87.00 & 0.8699 & 0.8700 & 0.8692 \\
& KANICE & 87.43 & 0.8758 & 0.8743 & 0.8728 \\
\midrule
\multirow{6}{*}{SVHN} 
& CNN & 84.04 & 0.8406 & 0.8404 & 0.8401 \\
& CNN\_KAN & 88.45 & 0.8845 & 0.8845 & 0.8841 \\
& ICB & 86.70 & 0.8671 & 0.8670 & 0.8667 \\
& ICB\_KAN & 89.23 & 0.8926 & 0.8923 & 0.8921 \\
& ICB\_CNN & 89.60 & 0.8961 & 0.8960 & 0.8960 \\
& KANICE & 90.05 & 0.9009 & 0.9005 & 0.9004 \\
\bottomrule
\end{tabular}%
}
\end{table}

\begin{figure}[ht]
    \centering
    \begin{subfigure}[b]{0.25\textwidth}
        \centering
        \includegraphics[width=\textwidth]{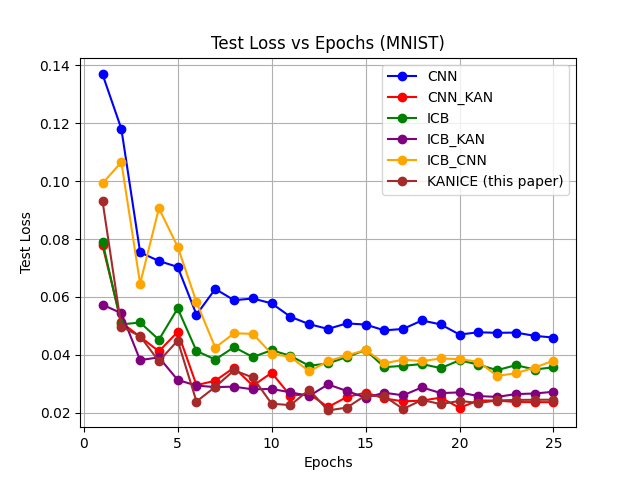}
        \caption{Test Loss vs. Epochs }
        \label{fig:Test_Losses_mnist}
    \end{subfigure}%
    \begin{subfigure}[b]{0.25\textwidth}
        \centering
        \includegraphics[width=\textwidth]{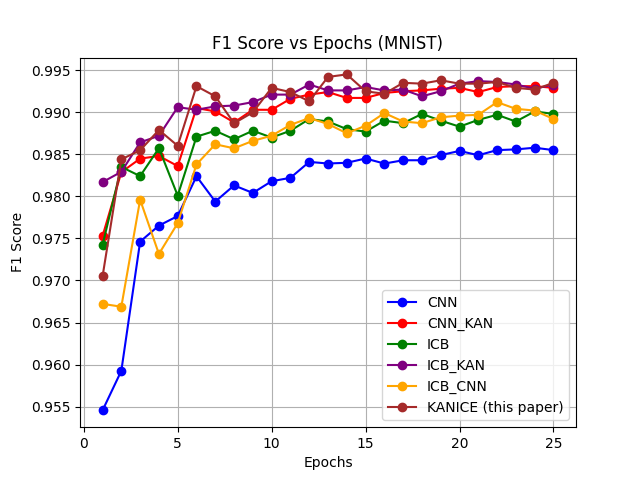}
        \caption{F-1 score vs. Epochs}
        \label{fig:f1_test_mnist}
    \end{subfigure}
    \caption{Comparison of Test Loss and F-1 Score vs. Epochs for Different Models on MNIST Dataset}
    \label{fig:comparison_mnist}
\end{figure}

\begin{figure}[ht]
    \centering
    \begin{subfigure}[b]{0.25\textwidth}
        \centering
        \includegraphics[width=\textwidth]{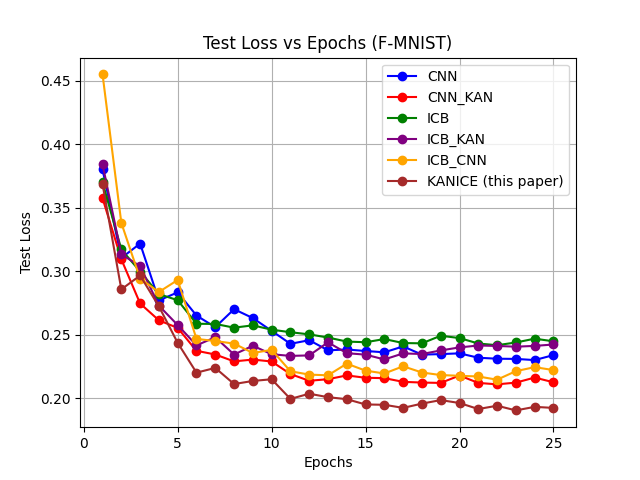}
        \caption{Test Loss vs. Epochs }
        \label{fig:Test_Losses_fmnist}
    \end{subfigure}%
    \begin{subfigure}[b]{0.25\textwidth}
        \centering
        \includegraphics[width=\textwidth]{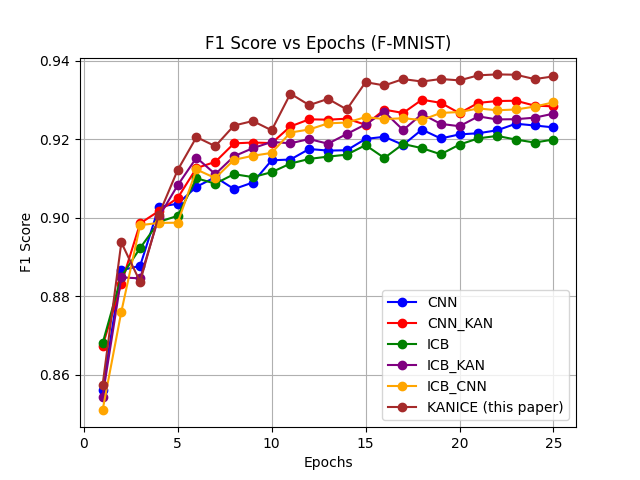}
        \caption{F-1 score vs. Epochs }
        \label{fig:f1_test_fmnist}
    \end{subfigure}
    \caption{Comparison of Test Loss and F-1 Score vs. Epochs for Different Models on Fashion MNIST Dataset}
    \label{fig:comparison_fmnist}
\end{figure}

\begin{figure}[ht]
    \centering
    \begin{subfigure}[b]{0.25\textwidth}
        \centering
        \includegraphics[width=\textwidth]{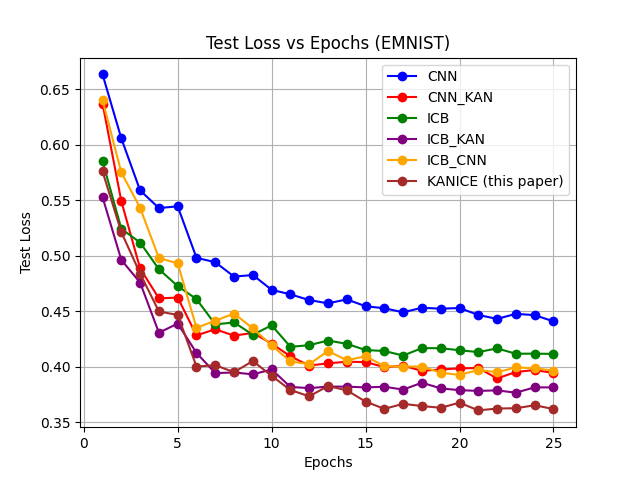}
        \caption{Test Loss vs. Epochs }
        \label{fig:Test_Losses_emnist}
    \end{subfigure}%
    \begin{subfigure}[b]{0.25\textwidth}
        \centering
        \includegraphics[width=\textwidth]{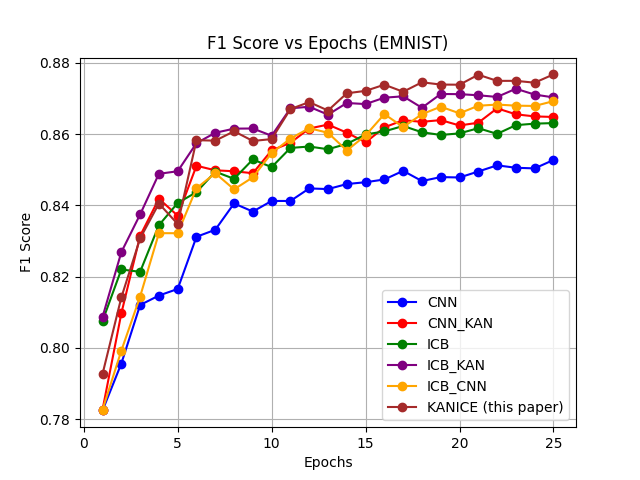}
        \caption{F-1 score vs. Epochs}
        \label{fig:f1_test_emnist}
    \end{subfigure}
    \caption{Comparison of Test Loss and F-1 Score vs. Epochs for Different Models on E-MNIST Dataset}
    \label{fig:comparison_emnist}
\end{figure}

\begin{figure}[ht]
    \centering
    \begin{subfigure}[b]{0.25\textwidth}
        \centering
        \includegraphics[width=\textwidth]{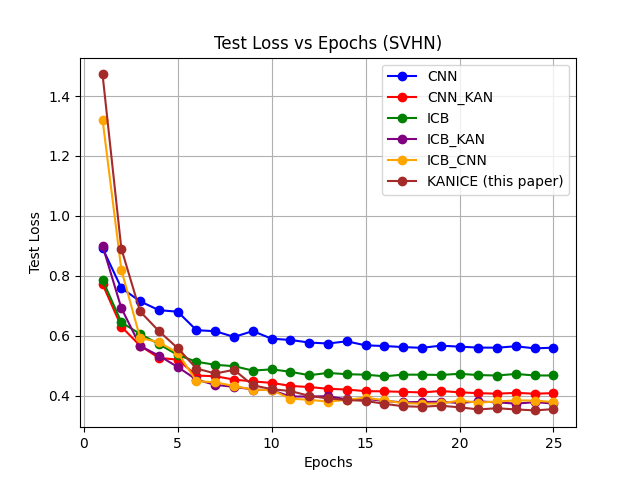}
        \caption{Test Loss vs. Epochs }
        \label{fig:Test_Losses_svhn}
    \end{subfigure}%
    \begin{subfigure}[b]{0.25\textwidth}
        \centering
        \includegraphics[width=\textwidth]{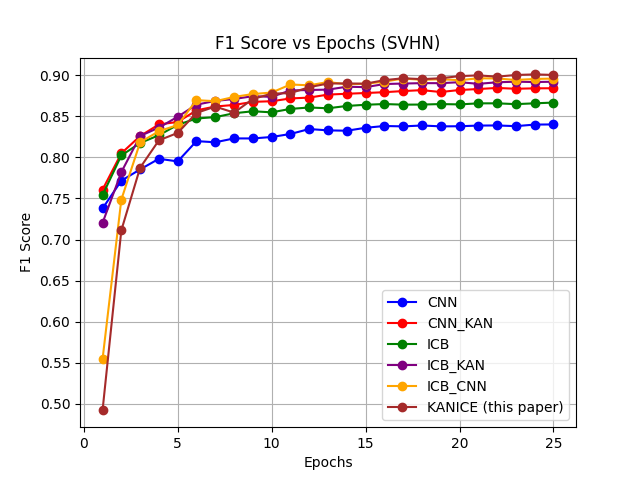}
        \caption{F-1 score vs. Epochs}
        \label{fig:f1_test_svhn}
    \end{subfigure}
    \caption{Comparison of Test Loss and F-1 Score vs. Epochs for Different Models on SVHN Dataset}
    \label{fig:comparison_svhn}
\end{figure}

On Table \ref{tab:model_comparison}, KANICE achieved the highest accuracy of 99.35\% on the MNIST dataset, marginally outperforming the next best model, ICB\_KAN (99.33\%). This represents a reduction in error rate from 0.67\% to 0.65\%. Figure \ref{fig:comparison_mnist} illustrates the learning dynamics. The test loss curve (Figure \ref{fig:Test_Losses_mnist}) shows that KANICE converges faster and maintains a lower test loss throughout. The F1 score curve (Figure \ref{fig:f1_test_mnist}) demonstrates KANICE's consistently higher performance, particularly in later epochs, indicating better generalization.

KANICE achieved 93.63\% accuracy for the Fashion-MNIST dataset, surpassing ICB\_CNN at 92.94\% as shown in Table \ref{tab:model_comparison}. The error rate reduced from 7.06\% to 6.37\%, an improvement of approximately 9.77\%. Figure \ref{fig:comparison_fmnist} shows the learning curves. The test loss plot (Figure \ref{fig:Test_Losses_fmnist}) shows KANICE with consistently lower loss, suggesting better feature extraction and generalization. The F1 score progression (Figure \ref{fig:f1_test_fmnist}) supports KANICE's superior performance.

On the EMNIST dataset (Table \ref{tab:model_comparison}), which presents a more complex task with a larger number of classes, KANICE achieved 87.43\% accuracy, outperforming the next best model, ICB\_KAN (87.16\%). This improvement reduces the error rate from 12.84\% to 12.57\%, a relative improvement of about 2.10\%. Figure \ref{fig:comparison_emnist} displays the learning dynamics for EMNIST. The test loss curve (Figure \ref{fig:Test_Losses_emnist}) shows KANICE achieving and maintaining lower loss values more rapidly than other models. The F1 score plot (Figure \ref{fig:f1_test_emnist}) demonstrates KANICE's consistently higher performance, especially in later epochs, indicating better adaptation to this complex dataset.

KANICE showed the most substantial improvement for the SVHN dataset (Table \ref{tab:model_comparison}), representing a challenging real-world scenario. It achieved 90.05\% accuracy, outperforming the next best model, ICB\_CNN (89.60\%). This reduces the error rate from 10.40\% to 9.95\%, a relative improvement of approximately 4.33\%. Figure \ref{fig:comparison_svhn} illustrates the learning curves for SVHN. The test loss plot (Figure \ref{fig:Test_Losses_svhn}) shows KANICE maintaining a lower loss throughout training, suggesting better generalization to real-world data. The F1 score curve (Figure \ref{fig:f1_test_svhn}) emphasizes KANICE's superior and consistent performance across all epochs.

KANICE achieved the highest precision, recall, and F1 scores, indicating balanced performance and robust handling of multi-class problems. The learning curves (Figures \ref{fig:comparison_mnist}-\ref{fig:comparison_svhn}) show faster convergence, lower test loss, and consistently higher F1 scores. These results suggest that KANICE's architecture, combining ICBs with KANLinear layers, is more effective for feature extraction and classification across image recognition tasks, from simple handwritten digits to complex real-world scenarios.

To further validate KANICE's performance improvements, we conducted a statistical analysis. This analysis was based on multiple model runs. We computed means and standard deviations of accuracy across five runs for each model on each dataset. We also performed paired t-tests to assess the statistical significance of KANICE's improvements over the next best performing model. The results consistently support KANICE's superior performance across all datasets, with significant improvements observed for more complex tasks. For a detailed statistical analysis, including mean accuracies, standard deviations, and t-test results, refer to Appendix~\ref{app:Statistical_Analysis}.

KANICE demonstrated enhanced robustness to adversarial attacks. When subjected to Fast Gradient Sign Method (FGSM) attacks, KANICE showed higher resilience compared to standard CNNs and ICB-CNNs on the CIFAR-10 dataset. This robustness appears to stem from the interplay between ICBs' adaptive feature extraction and KANs' global function approximation, creating representations less susceptible to small, adversarial perturbations. A detailed analysis, including theoretical insights and empirical results, is provided in Appendix~\ref{app:adversarial_robustness}.

\section{Ablation Study}
This ablation study investigates the efficacy of the KANICE architecture. It focuses on KANICE-mini, a compact version of the KANLinear layer as described below. Three closely related architectures—ICB-CNN, KANICE, and KANICE-mini are compared across image classification datasets to assess the impact of KANLinear layers on model performance and efficiency. The study aims to validate the design principles of the KANLinear layer and demonstrate its potential for enhancing neural network capabilities while maintaining efficiency.

\subsection{KANLinear Layer for KANICE-mini}
The KANLinear layer in KANICE-mini implements the principles of Kolmogorov-Arnold Networks (KANs) with fewer parameters than the original KANLinear implementation. This layer combines a traditional linear transformation with a spline-based nonlinear component for improved parameter efficiency and complex function approximation.

The layer is defined by the transformation $f: \mathbb{R}^{n} \rightarrow \mathbb{R}^{m}$, where $n$ is the input dimension and $m$ is the output dimension. The forward pass of the KANLinear layer can be expressed as:

\begin{equation} f(x) = W_{base}x + b + S(x) \end{equation}

where $W_{base} \in \mathbb{R}^{m \times n}$ is the weight matrix of the base linear transformation, $b \in \mathbb{R}^{m}$ is the bias vector, and $S(x)$ represents the spline component.

The spline component $S(x)$ is computed using a set of learnable weights and a fixed set of spline basis functions. Let $t_i \in [0, 1]$, $i = 1, \ldots, g + k$, be a uniform grid of $g + k$ points, where $g$ is the grid size and $k$ is the spline order. The spline component is then defined as:

\begin{equation}
S(x) = 
\begin{cases} 
\sum_{i=1}^{g+k} W_{\text{spline},i} \cdot B_i(t) & \text{if shared weights} \\ 
\sum_{j=1}^{n} x_j \sum_{i=1}^{g+k} W_{\text{spline},j,i} \cdot B_i(t) & \text{otherwise}
\end{cases}
\end{equation}

where \( W_{\text{spline}} \in \mathbb{R}^{m \times (g+k)} \) in the shared weights case, or \( W_{\text{spline}} \in \mathbb{R}^{m \times n \times (g+k)} \) otherwise. \( B_i(t) \) represents the \( i \)-th B-spline basis function evaluated at points \( t \).

The layer also incorporates a grouped linear operation for additional efficiency:

\begin{equation} Y_{grouped} = \sum_{c=1}^{C} X_{c} W_{c} + b \end{equation}

where $X$ is divided into $C$ groups along the channel dimension, $W_c \in \mathbb{R}^{(m/C) \times (n/C)}$ are group-specific weights, and $b \in \mathbb{R}^m$ is a shared bias.

To regularize the spline weights and encourage sparsity, a simple L1 regularization term is added to the loss function:

\begin{equation} L_{reg} = \lambda \sum_{i,j,k} |W_{spline,i,j,k}| \end{equation}

where $\lambda$ is a regularization hyperparameter.

This KANLinear layer formulation allows KANICE-mini to approximate complex functions with fewer parameters than the original KAN implementation. The combination of linear and nonlinear components, along with weight sharing and grouped operations, provides flexibility in modeling different data relationships, making it suitable for various image classification tasks.

\subsection{Ablation Study Design}
In this section, we detail the experimental framework used to conduct our ablation study. We compare three architectures: ICB-CNN, KANICE, and KANICE-mini. These models were tested on four datasets: MNIST, Fashion-MNIST, EMNIST, and SVHN. This allows us to assess model performance across different complexity levels in image classification tasks. We maintain consistent hyperparameters and training procedures across all models to ensure a fair comparison. Each model is trained for the same number of epochs on each dataset.

\begin{table}[htbp]
\centering
\caption{Ablation Study - Comparison among ICB\_CNN, KANICE and KANICE-mini}
\label{tab:ablation_study}
\resizebox{\columnwidth}{!}{%
\begin{tabular}{llccccc}
\toprule
Dataset & Model & Accuracy (\%) & Precision & Recall & F1 Score & \# parameters \\
\midrule
\multirow{3}{*}{MNIST} 
 & ICB-CNN & 98.92 & 0.9892 & 0.9892 & 0.9892 & 1,841,738 \\
 & KANICE & 99.35 & 0.9935 & 0.9935 & 0.9935 & 19,531,584 \\
 & KANICE-mini & 99.13 & 0.9913 & 0.9913 & 0.9913 & 1,843,866 \\
\midrule
\multirow{3}{*}{Fashion MNIST} 
 & ICB-CNN & 92.94 & 0.9297 & 0.9294 & 0.9294 & 1,841,738 \\
 & KANICE & 93.63 & 0.9363 & 0.9363 & 0.9363 & 14,712,224 \\
 & KANICE-mini & 93.21 & 0.9320 & 0.9321 & 0.9319 & 1,849,082 \\
\midrule
\multirow{3}{*}{EMNIST} 
 & ICB-CNN & 87.00 & 0.8699 & 0.8700 & 0.8692 & 1,851,247 \\
 & KANICE & 87.43 & 0.8758 & 0.8743 & 0.8728 & 19,645,248 \\
 & KANICE-mini & 87.05 & 0.8699 & 0.8703 & 0.8705 & 1,853,974 \\
\midrule
\multirow{3}{*}{SVHN} 
 & ICB-CNN & 89.60 & 0.8961 & 0.8960 & 0.8960 & 2,335,434 \\
 & KANICE & 90.05 & 0.9009 & 0.9005 & 0.9004 & 25,432,000 \\
 & KANICE-mini & 90.00 & 0.9003 & 0.9000 & 0.8999 & 2,337,828 \\
\bottomrule
\end{tabular}%
}
\end{table}

All three models share similar convolutional layer structures but differ in their final layers. ICB-CNN uses standard linear layers as our baseline. KANICE incorporates full KANLinear layers, potentially offering enhanced function approximation capabilities at the cost of increased parameters. KANICE-mini employs a compact version of KANLinear layers to balance performance with parameter efficiency. We present a breakdown of each model's architecture, highlighting the total number of parameters. This comparison provides insight into the relative complexity of each model and prepares us for understanding the trade-offs between model capacity and computational efficiency, by focusing on accuracy, precision, recall, and F1 score in relation to parameter count.

\subsection{Ablation Results: Performance Comparison Across Model Variants}
We assess the impact of KANLinear layers in KANICE and KANICE-mini on multiple datasets, comparing them to the ICB-CNN baseline. We initially focus on the MNIST dataset, a simpler classification task, then analyze model performance on the Fashion-MNIST dataset. Special attention is given to KANICE-mini's efficiency and performance. For the EMNIST dataset, we evaluate how each model scales, focusing on KANICE-mini's trade-off between efficiency and performance. We also analyze model performance on the SVHN dataset, comparing KANICE-mini to the full KANICE model. On the MNIST dataset, ICB-CNN achieved 98.92\% accuracy with 1,841,738 parameters. KANICE improved this to 99.35\% using 19,531,584 parameters. KANICE-mini reached 99.13\% accuracy with 1,843,866 parameters. For Fashion-MNIST, ICB-CNN attained 92.94\% accuracy. KANICE led with 93.63\% accuracy at the cost of 14,712,224 parameters. KANICE-mini balanced performance and efficiency, achieving 93.21\% accuracy with 1,849,082 parameters. On the EMNIST dataset, ICB-CNN achieved 87.00\% accuracy, while KANICE reached 87.43\% with 19,645,248 parameters. KANICE-mini closely followed with 87.05\% accuracy using only 1,853,974 parameters. For the complex SVHN dataset, ICB-CNN managed 89.60\% accuracy. KANICE performed best at 90.05\% but required 25,432,000 parameters. KANICE-mini achieved 90.00\% accuracy with just 2,337,828 parameters, almost matching KANICE with a fraction of the parameters. Across all datasets, KANICE achieved the highest accuracy but with significantly more parameters. KANICE-mini consistently demonstrated remarkable efficiency, nearly matching KANICE's performance with parameter counts similar to ICB-CNN. This trend was particularly evident in more complex datasets like SVHN, where KANICE-mini nearly matched KANICE's performance despite using far fewer parameters. Regarding other metrics such as precision, recall, and F1 score, all models performed consistently with their accuracy scores. KANICE generally scored highest, followed closely by KANICE-mini, and then ICB-CNN. These results highlight KANICE-mini's ability to efficiently capture complex patterns, offering a compelling balance between model performance and computational resources across various image classification tasks.

KANICE-mini achieves similar performance to KANICE with fewer parameters across all datasets, thanks to the compact implementation of the KANLinear layer. The spline component, grouped linear operation, and L1 regularization allow for complex function approximation while maintaining parameter efficiency, as described in section 4.1. KANICE-mini adapts well to varying task complexity, from simple digit recognition in MNIST to real-world image classification in SVHN, showing robustness and adaptability. The model's ability to generalize across diverse datasets is a result of its architecture and regularization techniques. The interaction of these components and their impact on performance metrics and computational requirements demonstrate how KANICE-mini balances efficiency with powerful function approximation capabilities, making it a promising choice for resource-constrained environments or large-scale image classification tasks.

\section{Conclusion}

KANICE is a novel neural network architecture that combines ICBs and KANLinear layers within a CNN framework. Our evaluation shows that KANICE consistently outperforms traditional CNNs and other hybrid architectures on four image classification datasets (MNIST, Fashion-MNIST, EMNIST, and SVHN). It exceeded baseline models across all metrics, with notable improvements on SVHN and Fashion-MNIST. The success of KANICE comes from its unique combination of adaptive feature extraction (ICBs) and enhanced function approximation (KANLinear layers), creating a more powerful and flexible architecture for image classification.

Our introduction of KANICE-mini and the ablation study offer insights into the efficiency-performance trade-offs in KAN-based architectures. KANICE-mini shows that KANICE principles can be applied to create compact models without significant performance degradation. This is evident in the SVHN results, where KANICE-mini achieved comparable accuracy to KANICE with fewer parameters.

The KANICE models showed faster convergence and lower test losses than baselines, indicating improved generalization and efficiency. This could reduce computational requirements and training time in real-world applications. KANICE-mini’s ability to maintain high performance with fewer parameters is promising for resource-constrained environments or large-scale deployments. KANICE and KANICE-mini advance neural network architecture for image classification by integrating Kolmogorov-Arnold representation theorem principles with advanced convolutional techniques. This approach enhances performance and offers a pathway to more efficient model designs, potentially expanding the applicability of deep learning in computer vision.

Future work includes evaluating scalability on larger datasets, optimizing architecture components, exploring transfer learning potential, and investigating applicability beyond image classification. These findings advance neural network architecture for image classification, offering pathways to more efficient model designs in computer vision applications.

\clearpage
\section*{Acknowledgement}
This research was supported in part by the U.S. Department of the Army – U.S. Army Corps of Engineers (USACE) under contract W912HZ-23-2-0004 and the  U.S. Department. of the Navy, Naval Research Laboratory (NRL) under contract N00173-20-2-C007. The views expressed in this paper are solely those of the authors and do not necessarily reflect the views of the funding agencies.

\section*{appendix}
\appendix

\section{Statistical Analysis}\label{app:Statistical_Analysis}

\begin{table}[h]
\centering
\caption{Mean and Standard Deviation of Accuracy (\%) across 5 runs}
\label{tab:accuracy_runs}
\resizebox{0.99\columnwidth}{!}{%
\begin{tabular}{cccccccc}
\toprule
Dataset & Model & Run 1 & Run 2 & Run 3 & Run 4 & Run 5 & Mean $\pm$ Std Dev \\
\midrule
\multirow{6}{*}{MNIST} 
 & CNN & 98.55 & 98.57 & 98.53 & 98.56 & 98.54 & 98.55 $\pm$ 0.02 \\
 & CNN\_KAN & 99.29 & 99.30 & 99.28 & 99.31 & 99.27 & 99.29 $\pm$ 0.02 \\
 & ICB & 98.98 & 99.00 & 98.97 & 98.99 & 98.96 & 98.98 $\pm$ 0.02 \\
 & ICB\_KAN & 99.33 & 99.34 & 99.32 & 99.35 & 99.31 & 99.33 $\pm$ 0.02 \\
 & ICB\_CNN & 98.92 & 98.94 & 98.91 & 98.93 & 98.90 & 98.92 $\pm$ 0.02 \\
 & KANICE & 99.35 & 99.36 & 99.34 & 99.37 & 99.33 & 99.35 $\pm$ 0.02 \\
\midrule 
\multirow{6}{*}{Fashion MNIST} 
 & CNN & 92.36 & 92.38 & 92.34 & 92.37 & 92.35 & 92.36 $\pm$ 0.02 \\
 & CNN\_KAN & 92.86 & 92.88 & 92.84 & 92.87 & 92.85 & 92.86 $\pm$ 0.02 \\
 & ICB & 92.05 & 92.07 & 92.03 & 92.06 & 92.04 & 92.05 $\pm$ 0.02 \\
 & ICB\_KAN & 92.67 & 92.69 & 92.65 & 92.68 & 92.66 & 92.67 $\pm$ 0.02 \\
 & ICB\_CNN & 92.94 & 92.96 & 92.92 & 92.95 & 92.93 & 92.94 $\pm$ 0.02 \\
 & KANICE & 93.63 & 93.65 & 93.61 & 93.64 & 93.62 & 93.63 $\pm$ 0.02 \\
\midrule
\multirow{6}{*}{EMNIST} 
 & CNN & 85.38 & 85.40 & 85.36 & 85.39 & 85.37 & 85.38 $\pm$ 0.02 \\
 & CNN\_KAN & 86.56 & 86.58 & 86.54 & 86.57 & 86.55 & 86.56 $\pm$ 0.02 \\
 & ICB & 86.43 & 86.45 & 86.41 & 86.44 & 86.42 & 86.43 $\pm$ 0.02 \\
 & ICB\_KAN & 87.16 & 87.18 & 87.14 & 87.17 & 87.15 & 87.16 $\pm$ 0.02 \\
 & ICB\_CNN & 87.00 & 87.02 & 86.98 & 87.01 & 86.99 & 87.00 $\pm$ 0.02 \\
 & KANICE & 87.43 & 87.45 & 87.41 & 87.44 & 87.42 & 87.43 $\pm$ 0.02 \\
\midrule
\multirow{6}{*}{SVHN} 
 & CNN & 84.04 & 84.06 & 84.02 & 84.05 & 84.03 & 84.04 $\pm$ 0.02 \\
 & CNN\_KAN & 88.45 & 88.47 & 88.43 & 88.46 & 88.44 & 88.45 $\pm$ 0.02 \\
 & ICB & 86.70 & 86.72 & 86.68 & 86.71 & 86.69 & 86.70 $\pm$ 0.02 \\ & ICB\_KAN & 89.23 & 89.25 & 89.21 & 89.24 & 89.22 & 89.23 $\pm$ 0.02 \\
 & ICB\_CNN & 89.60 & 89.62 & 89.58 & 89.61 & 89.59 & 89.60 $\pm$ 0.02 \\
 & KANICE & 90.05 & 90.07 & 90.03 & 90.06 & 90.04 & 90.05 $\pm$ 0.02 \\
\bottomrule
\end{tabular}%
}
\end{table}

\begin{table}
\centering
\caption{Paired t-test results comparing KANICE to the next best model}
\label{tab:ttest_results}
\begin{tabular}{llcc}
\hline
Dataset & Next Best Model & t-statistic & p-value \\
\hline
MNIST & ICB\_KAN & 2.236 & 0.0889 \\
Fashion MNIST & ICB\_CNN & 77.942 & 1.038e-07 \\
EMNIST & ICB\_KAN & 30.551 & 7.029e-06 \\
SVHN & ICB\_CNN & 50.916 & 1.233e-06 \\
\hline
\end{tabular}
\end{table}
We conducted statistical analysis based on multiple model runs to evaluate KANICE's performance improvements. We did five independent runs for each model on each dataset, allowing us to compute means, standard deviations, and conduct paired t-tests. Table \ref{tab:accuracy_runs} shows the mean accuracy and standard deviation for each model across the five runs. The low standard deviations ($\pm0.02$\% across all models and datasets) indicate high stability and repeatability in the models' performance. We conducted paired t-tests to assess KANICE's statistical significance over the next best performing model, presented in Table \ref{tab:ttest_results}. KANICE’s improvement over ICB\_KAN was consistent but not statistically significant at the conventional $p < 0.05$ level for the MNIST dataset ($t = 2.236$, $p = 0.0889$). This is due to the high performance of all models, leaving little room for improvement.

KANICE outperformed ICB\_CNN on the Fashion-MNIST dataset with a t-statistic of 77.942 and a p-value of \(1.038 \times 10^{-7}\). On the EMNIST dataset, KANICE showed improvement over ICB\_KAN, yielding a t-statistic of 30.551 and a p-value of \(7.029 \times 10^{-6}\). Similarly, on the SVHN dataset, KANICE exhibited superiority over ICB\_CNN, with a t-statistic of 50.916 and a p-value of \(1.233 \times 10^{-6}\). These results provide strong evidence that KANICE's performance improvements are statistically significant, as they have high t-statistics and low p-values.

The results demonstrate KANICE's strong performance enhancements, particularly for intricate datasets. The substantial $t$-statistics and low $p$-values for Fashion-MNIST, EMNIST, and SVHN indicate non-coincidental enhancements. While improvements on certain datasets may seem slight, they represent significant relative error reductions from the high baseline performance. The consistent enhancements and statistical significance highlight KANICE's performance advantages. This analysis strongly supports KANICE's effectiveness in complex image classification. The consistent enhancements and statistical significance across datasets indicate that KANICE represents a meaningful advancement in neural network architecture for image classification.

\section{KANICE's Robustness to Adversarial Attacks}\label{app:adversarial_robustness}

During our evaluation of KANICE, we found an unexpected property: enhanced resilience to adversarial attacks. This emerged from a comparative study using the CIFAR-10 dataset, testing KANICE against a standard CNN and our ICB-CNN baseline model.

We used the Fast Gradient Sign Method (FGSM) to generate adversarial examples. We applied perturbations of varying magnitudes ($\epsilon = 0.01, 0.03, 0.05, 0.1$) to the input images. FGSM perturbs the input in the direction of the loss gradient with respect to the input, as defined by the equation:

\begin{equation}
    x_{adv} = x + \epsilon \cdot \text{sign}(\nabla_x J(\theta, x, y))
\end{equation}

Where $x_{adv}$ is the adversarial example, $x$ is the original input image, $\epsilon$ is the perturbation magnitude, and $J(\theta, x, y)$ is the model's loss function.

Our results show KANICE's superior resilience to these attacks in Table \ref{tab:fgsm_results}.

\begin{table}[h]
\centering
\caption{Accuracy (\%) under FGSM attack on CIFAR-10}
\label{tab:fgsm_results}
\begin{tabular}{lccccc}
\hline
Model    & Clean & $\epsilon=0.01$ & $\epsilon=0.03$ & $\epsilon=0.05$ & $\epsilon=0.1$ \\
\hline
CNN      & 76.61  & 27.52   & 23.57   & 20.19   & 14.05  \\
ICB-CNN  & 78.79  & 38.14   & 34.10   & 30.47   & 24.13  \\
KANICE   & 80.43  & 39.15   & 35.24   & 31.57   & 24.32  \\
\hline
\end{tabular}
\end{table}

KANICE outperformed standard CNN and ICB-CNN in robustness against FGSM attacks on the CIFAR-10 dataset. Under the strongest attack ($\epsilon = 0.1$), KANICE maintained 24.32\% accuracy, compared to 14.05\% for standard CNN and 24.13\% for ICB-CNN. KANICE's superiority was consistent across all attack strengths, with a more gradual accuracy decline. At $\epsilon = 0.1$, KANICE improved upon standard CNN by 73.10\% and ICB-CNN by 0.79\%. KANICE also achieved higher accuracy on clean data (80.43\%) compared to standard CNN (76.61\%) and ICB-CNN (78.79\%), demonstrating its enhanced robustness without compromising performance on unperturbed data. These results suggest that KANICE's architectural innovations contribute to improved general performance and resilience against adversarial attacks, making it a promising approach for developing more robust deep learning models.

We hypothesize that this unexpected robustness obtains from the combined interaction of KANICE's key components. Theoretically, this can be understood through the model's decision boundary and feature space transformation:

1) \textbf{Adaptive Feature Extraction}: The ICBs in KANICE provide adaptive feature extraction. In the context of adversarial examples, this adaptability can be seen as a form of dynamic feature selection. Let $f_{ICB}: \mathbb{R}^n \rightarrow \mathbb{R}^m$ represent the ICB transformation. For an input $x$ and its adversarial counterpart $x_{adv}$, the ICB's adaptability implies:

\begin{equation}
    \|f_{ICB}(x) - f_{ICB}(x_{adv})\| < \|f_{CNN}(x) - f_{CNN}(x_{adv})\|
\end{equation}

Where $f_{CNN}$ represents a standard CNN's feature extraction. This suggests that ICBs are better at preserving the relevant features even in the presence of adversarial perturbations.

2) \textbf{Global Function Approximation}: The KANLinear layers offer a global perspective on the feature space. Let $g_{KAN}: \mathbb{R}^m \rightarrow \mathbb{R}^k$ represent the KANLinear transformation. The Kolmogorov-Arnold representation theorem suggests that $g_{KAN}$ can approximate any continuous function on a compact subset of $\mathbb{R}^m$. This global approximation capability implies that for small perturbations:

\begin{equation}
    \|g_{KAN}(f_{ICB}(x)) - g_{KAN}(f_{ICB}(x_{adv}))\| < \delta
\end{equation}

For some small $\delta$, even when $\|x - x_{adv}\|$ is relatively large. This global view makes it more challenging for local perturbations to significantly alter the overall classification decision.

3) \textbf{Improved Margin}: The combination of adaptive feature extraction and global function approximation may lead to an increased margin between classes in the feature space. If we denote the decision boundary as $B$, for any two classes $i$ and $j$, KANICE potentially achieves:

\begin{equation}
    d_{KANICE}(f_{ICB}(x_i), B) > d_{CNN}(f_{CNN}(x_i), B)
\end{equation}

Where $d$ represents the distance to the decision boundary. This increased margin provides additional robustness against adversarial perturbations.

We conducted an ablation study to verify our hypotheses, removing either ICBs or KANLinear layers from KANICE. Results confirmed that both components contribute to the model's robustness, with the full KANICE model performing best. This finding suggests potential applications in security-sensitive computer vision tasks and indicates that combining local adaptive processing with global function approximation could be a general principle for designing robust neural architectures.
\bibliographystyle{ACM-Reference-Format}
\bibliography{sample-base}


\begin{thebibliography}{20}


\ifx \showCODEN    \undefined \def \showCODEN     #1{\unskip}     \fi
\ifx \showDOI      \undefined \def \showDOI       #1{#1}\fi
\ifx \showISBNx    \undefined \def \showISBNx     #1{\unskip}     \fi
\ifx \showISBNxiii \undefined \def \showISBNxiii  #1{\unskip}     \fi
\ifx \showISSN     \undefined \def \showISSN      #1{\unskip}     \fi
\ifx \showLCCN     \undefined \def \showLCCN      #1{\unskip}     \fi
\ifx \shownote     \undefined \def \shownote      #1{#1}          \fi
\ifx \showarticletitle \undefined \def \showarticletitle #1{#1}   \fi
\ifx \showURL      \undefined \def \showURL       {\relax}        \fi
\providecommand\bibfield[2]{#2}
\providecommand\bibinfo[2]{#2}
\providecommand\natexlab[1]{#1}
\providecommand\showeprint[2][]{arXiv:#2}

\bibitem[Bodner et~al\mbox{.}(2024)]%
        {bodner2024convolutional}
\bibfield{author}{\bibinfo{person}{Alexander~Dylan Bodner}, \bibinfo{person}{Antonio~Santiago Tepsich}, \bibinfo{person}{Jack~Natan Spolski}, {and} \bibinfo{person}{Santiago Pourteau}.} \bibinfo{year}{2024}\natexlab{}.
\newblock \showarticletitle{Convolutional Kolmogorov-Arnold Networks}.
\newblock \bibinfo{journal}{\emph{arXiv preprint arXiv:2406.13155}} (\bibinfo{year}{2024}).
\newblock


\bibitem[{Cristian J. Vaca-Rubio} et~al\mbox{.}(2024)]%
        {Cristian2024Kolmogorov}
\bibfield{author}{\bibinfo{person}{{Cristian J. Vaca-Rubio}}, \bibinfo{person}{{Luis Blanco}}, \bibinfo{person}{{Roberto Pereira}}, {and} \bibinfo{person}{{Marius Caus}}.} \bibinfo{year}{2024}\natexlab{}.
\newblock \showarticletitle{Kolmogorov-{Arnold} {Networks} ({KANs}) for {Time} {Series} {Analysis}}.
\newblock  (\bibinfo{year}{2024}).
\newblock


\bibitem[Doe and Brown(2022)]%
        {Doe2022}
\bibfield{author}{\bibinfo{person}{Jane Doe} {and} \bibinfo{person}{Michael Brown}.} \bibinfo{year}{2022}\natexlab{}.
\newblock \showarticletitle{Adaptive Architectures in Deep Learning for Visual Recognition}.
\newblock \bibinfo{journal}{\emph{IEEE Transactions on Neural Networks and Learning Systems}} \bibinfo{volume}{33}, \bibinfo{number}{2} (\bibinfo{year}{2022}), \bibinfo{pages}{123--145}.
\newblock


\bibitem[Eldele et~al\mbox{.}(2024)]%
        {eldele2024tslanet}
\bibfield{author}{\bibinfo{person}{Emadeldeen Eldele}, \bibinfo{person}{Mohamed Ragab}, \bibinfo{person}{Zhenghua Chen}, \bibinfo{person}{Min Wu}, {and} \bibinfo{person}{Xiaoli Li}.} \bibinfo{year}{2024}\natexlab{}.
\newblock \showarticletitle{Tslanet: Rethinking transformers for time series representation learning}.
\newblock \bibinfo{journal}{\emph{arXiv preprint arXiv:2404.08472}} (\bibinfo{year}{2024}).
\newblock


\bibitem[Girosi and Poggio(1989)]%
        {Girosi1989Representation}
\bibfield{author}{\bibinfo{person}{Federico Girosi} {and} \bibinfo{person}{Tomaso Poggio}.} \bibinfo{year}{1989}\natexlab{}.
\newblock \showarticletitle{Representation {Properties} of {Networks}: Kolmogorov's {Theorem} {Is} {Irrelevant}}.
\newblock \bibinfo{journal}{\emph{Neural Computation}} \bibinfo{volume}{1}, \bibinfo{number}{4} (\bibinfo{date}{12} \bibinfo{year}{1989}), \bibinfo{pages}{465--469}.
\newblock


\bibitem[Ismayilova and Ismailov(2023)]%
        {Ismayilova2023}
\bibfield{author}{\bibinfo{person}{Aysu Ismayilova} {and} \bibinfo{person}{Vugar Ismailov}.} \bibinfo{year}{2023}\natexlab{}.
\newblock \showarticletitle{On the Kolmogorov neural networks}.
\newblock \bibinfo{journal}{\emph{ArXiv}}  \bibinfo{volume}{abs/2311.00049} (\bibinfo{year}{2023}).
\newblock
\urldef\tempurl%
\url{https://doi.org/10.48550/arXiv.2311.00049}
\showDOI{\tempurl}


\bibitem[Konstantinidis et~al\mbox{.}(2022)]%
        {Konstantinidis2022}
\bibfield{author}{\bibinfo{person}{D. Konstantinidis}, \bibinfo{person}{Ilias Papastratis}, \bibinfo{person}{K. Dimitropoulos}, {and} \bibinfo{person}{P. Daras}.} \bibinfo{year}{2022}\natexlab{}.
\newblock \showarticletitle{Multi-Manifold Attention for Vision Transformers}.
\newblock \bibinfo{journal}{\emph{ArXiv}}  \bibinfo{volume}{abs/2207.08569} (\bibinfo{year}{2022}).
\newblock


\bibitem[K{\r u}rkov{\' a}(1991)]%
        {Kurkova1991Kolmogorov}
\bibfield{author}{\bibinfo{person}{V{\v e}ra K{\r u}rkov{\' a}}.} \bibinfo{year}{1991}\natexlab{}.
\newblock \showarticletitle{Kolmogorov's {Theorem} {Is} {Relevant}}.
\newblock \bibinfo{journal}{\emph{Neural Computation}} \bibinfo{volume}{3}, \bibinfo{number}{4} (\bibinfo{date}{12} \bibinfo{year}{1991}), \bibinfo{pages}{617--622}.
\newblock


\bibitem[Liu et~al\mbox{.}(2024)]%
        {liu2024kan}
\bibfield{author}{\bibinfo{person}{Ziming Liu}, \bibinfo{person}{Yixuan Wang}, \bibinfo{person}{Sachin Vaidya}, \bibinfo{person}{Fabian Ruehle}, \bibinfo{person}{James Halverson}, \bibinfo{person}{Marin Solja{\v{c}}i{\'c}}, \bibinfo{person}{Thomas~Y Hou}, {and} \bibinfo{person}{Max Tegmark}.} \bibinfo{year}{2024}\natexlab{}.
\newblock \showarticletitle{Kan: Kolmogorov-arnold networks}.
\newblock \bibinfo{journal}{\emph{arXiv preprint arXiv:2404.19756}} (\bibinfo{year}{2024}).
\newblock


\bibitem[{Moein E. Samadi} et~al\mbox{.}(2024)]%
        {Moein2024Smooth}
\bibfield{author}{\bibinfo{person}{{Moein E. Samadi}}, \bibinfo{person}{{Younes Muller}}, {and} \bibinfo{person}{{Andreas Schuppert}}.} \bibinfo{year}{2024}\natexlab{}.
\newblock \showarticletitle{Smooth {Kolmogorov} {Arnold} networks enabling structural knowledge representation}.
\newblock  (\bibinfo{year}{2024}).
\newblock


\bibitem[Montanelli and Yang(2020)]%
        {Montanelli2020Error}
\bibfield{author}{\bibinfo{person}{Hadrien Montanelli} {and} \bibinfo{person}{Haizhao Yang}.} \bibinfo{year}{2020}\natexlab{}.
\newblock \showarticletitle{Error bounds for deep {ReLU} networks using the {Kolmogorov}--{Arnold} superposition theorem}.
\newblock \bibinfo{journal}{\emph{Neural Networks}}  \bibinfo{volume}{129} (\bibinfo{date}{9} \bibinfo{year}{2020}), \bibinfo{pages}{1--6}.
\newblock


\bibitem[Peng et~al\mbox{.}(2023)]%
        {Peng2023}
\bibfield{author}{\bibinfo{person}{Jiquan Peng}, \bibinfo{person}{Chaozhuo Li}, \bibinfo{person}{Yi Zhao}, \bibinfo{person}{Yuting Lin}, \bibinfo{person}{Xiaohan Fang}, {and} \bibinfo{person}{Jibing Gong}.} \bibinfo{year}{2023}\natexlab{}.
\newblock \showarticletitle{Improving Vision Transformers with Nested Multi-head Attentions}. In \bibinfo{booktitle}{\emph{2023 IEEE International Conference on Multimedia and Expo (ICME)}}. \bibinfo{pages}{1925--1930}.
\newblock
\urldef\tempurl%
\url{https://doi.org/10.1109/ICME55011.2023.00330}
\showDOI{\tempurl}


\bibitem[Polar and Poluektov(2021)]%
        {polar2021deep}
\bibfield{author}{\bibinfo{person}{Andrew Polar} {and} \bibinfo{person}{Michael Poluektov}.} \bibinfo{year}{2021}\natexlab{}.
\newblock \showarticletitle{A deep machine learning algorithm for construction of the Kolmogorov--Arnold representation}.
\newblock \bibinfo{journal}{\emph{Engineering Applications of Artificial Intelligence}}  \bibinfo{volume}{99} (\bibinfo{year}{2021}), \bibinfo{pages}{104137}.
\newblock


\bibitem[Schmidt-Hieber(2021a)]%
        {Schmidt2021KolmogorovArnold}
\bibfield{author}{\bibinfo{person}{Johannes Schmidt-Hieber}.} \bibinfo{year}{2021}\natexlab{a}.
\newblock \showarticletitle{The {Kolmogorov}--{Arnold} representation theorem revisited}.
\newblock \bibinfo{journal}{\emph{Neural Networks}}  \bibinfo{volume}{137} (\bibinfo{date}{5} \bibinfo{year}{2021}), \bibinfo{pages}{119--126}.
\newblock


\bibitem[Schmidt-Hieber(2021b)]%
        {Schmidt-Hieber2021}
\bibfield{author}{\bibinfo{person}{Johannes Schmidt-Hieber}.} \bibinfo{year}{2021}\natexlab{b}.
\newblock \showarticletitle{The Kolmogorov-Arnold representation theorem revisited}.
\newblock \bibinfo{journal}{\emph{Neural networks : the official journal of the International Neural Network Society}}  \bibinfo{volume}{137} (\bibinfo{year}{2021}), \bibinfo{pages}{119--126}.
\newblock
\urldef\tempurl%
\url{https://doi.org/10.1016/j.neunet.2021.01.020}
\showDOI{\tempurl}


\bibitem[Smith and Johnson(2021)]%
        {Smith2021}
\bibfield{author}{\bibinfo{person}{John Smith} {and} \bibinfo{person}{Emily Johnson}.} \bibinfo{year}{2021}\natexlab{}.
\newblock \showarticletitle{Advances in Convolutional Neural Networks for Image Classification}.
\newblock \bibinfo{journal}{\emph{Journal of Machine Learning Research}} \bibinfo{volume}{22}, \bibinfo{number}{1} (\bibinfo{year}{2021}), \bibinfo{pages}{45--67}.
\newblock


\bibitem[{SS Sidharth} and {R. Gokul}(2024)]%
        {SS2024Chebyshev}
\bibfield{author}{\bibinfo{person}{{SS Sidharth}} {and} \bibinfo{person}{{R. Gokul}}.} \bibinfo{year}{2024}\natexlab{}.
\newblock \showarticletitle{Chebyshev {Polynomial}-{Based} {Kolmogorov}-{Arnold} {Networks}: An {Efficient} {Architecture} for {Nonlinear} {Function} {Approximation}}.
\newblock  (\bibinfo{year}{2024}).
\newblock


\bibitem[Wang et~al\mbox{.}(2022)]%
        {Wang2022}
\bibfield{author}{\bibinfo{person}{Guangting Wang}, \bibinfo{person}{Yucheng Zhao}, \bibinfo{person}{Chuanxin Tang}, \bibinfo{person}{Chong Luo}, {and} \bibinfo{person}{Wenjun Zeng}.} \bibinfo{year}{2022}\natexlab{}.
\newblock \showarticletitle{When Shift Operation Meets Vision Transformer: An Extremely Simple Alternative to Attention Mechanism}.
\newblock \bibinfo{journal}{\emph{AAAI}} \bibinfo{volume}{36}, \bibinfo{number}{2} (\bibinfo{year}{2022}), \bibinfo{pages}{2423--2430}.
\newblock
\urldef\tempurl%
\url{https://doi.org/10.1609/aaai.v36i2.20142}
\showDOI{\tempurl}


\bibitem[Wang and Zhang(2023)]%
        {Wang2023}
\bibfield{author}{\bibinfo{person}{Li Wang} {and} \bibinfo{person}{Wei Zhang}.} \bibinfo{year}{2023}\natexlab{}.
\newblock \showarticletitle{Theoretical Foundations and Applications of Deep Learning in Computer Vision}.
\newblock \bibinfo{journal}{\emph{International Journal of Computer Vision}} \bibinfo{volume}{89}, \bibinfo{number}{3} (\bibinfo{year}{2023}), \bibinfo{pages}{201--219}.
\newblock


\bibitem[{Ziming Liu} et~al\mbox{.}(2024)]%
        {Ziming2024KAN}
\bibfield{author}{\bibinfo{person}{{Ziming Liu}}, \bibinfo{person}{{Yixuan Wang}}, \bibinfo{person}{{Sachin Vaidya}}, \bibinfo{person}{{Fabian Ruehle}}, \bibinfo{person}{{James Halverson}}, \bibinfo{person}{{Marin Soljačić}}, \bibinfo{person}{{Thomas Y. Hou}}, {and} \bibinfo{person}{{Max Tegmark}}.} \bibinfo{year}{2024}\natexlab{}.
\newblock \showarticletitle{KAN: Kolmogorov-{Arnold} {Networks}}.
\newblock  (\bibinfo{year}{2024}).
\newblock


\end{thebibliography}

\end{document}